\documentclass[arxiv]{custom}

\usepackage{multirow}

\conference{}

\usepackage{graphicx,etoolbox}
\usepackage{subfigure}
\usepackage{booktabs}
\usepackage{hyperref}
\usepackage{url}
\usepackage{algorithm}
\usepackage{algorithmic}
\usepackage{amsmath}
\usepackage{amssymb}
\usepackage{mathtools}
\usepackage{amsthm}
\usepackage{multirow}
\usepackage[bibliography=common,appendix=inline]{apxproof}

\newtheoremstyle{mystyle}
{}
{}
{\itshape}
{}
{\bfseries}
{.}
{ }
{}
\theoremstyle{mystyle}
\newtheoremrep{lemma}{Lemma}
\newtheoremrep{theorem}{Theorem}

\usepackage{amsthm} 
\usepackage{thmtools}

\declaretheoremstyle[
    headpunct={},
    postheadspace={0.75em},
    qed=\qedsymbol,
  spacebelow=0.5em
]{mystyle} 
\declaretheorem[name={{\it Proof}},style=mystyle,unnumbered,
]{prf}

\usepackage{bm}
\newcommand{\bA}{\mathbf{A}}
\newcommand{\bb}{\mathbf{b}}
\newcommand{\bc}{\mathbf{c}}
\newcommand{\bx}{\mathbf{x}}
\newcommand{\by}{\mathbf{y}}
\newcommand{\bz}{\mathbf{z}}
\newcommand{\bzl}{\mathbf{z}^{l}}
\newcommand{\bzu}{\mathbf{z}^{u}}
\newcommand{\bl}{\mathbf{l}}
\newcommand{\bu}{\mathbf{u}}

\newcommand{\pd}{\mathbf{p}^{\text{d}}}
\newcommand{\pg}{\mathbf{p}^{\text{g}}}
\newcommand{\pf}{\mathbf{p}^{\text{f}}}
\newcommand{\PTDF}{\text{PTDF}}

\newcommand{\pgmin}{\underline{\mathbf{p}}^{\text{g}}}
\newcommand{\pgmax}{\overline{\mathbf{p}}^{\text{g}}}
\newcommand{\pfmin}{\underline{\mathbf{p}}^{\text{f}}}
\newcommand{\pfmax}{\overline{\mathbf{p}}^{\text{f}}}

\DeclareMathSymbol{\shortminus}{\mathbin}{AMSa}{"39}

\title{\titlesize Dual Interior Point Optimization Learning}

\newcommand{\nsfai}{{\fontsize{11}{12}\selectfont\normalfont NSF AI Institute for Advances in Optimization}}
\newcommand{\gatech}{{\fontsize{11}{12}\selectfont\normalfont Georgia Institute of Technology, Atlanta, USA}}
\author{
Michael Klamkin \quad Mathieu Tanneau \quad Pascal Van Hentenryck\\[0.15em]\nsfai\\\gatech
}
\authorlist{Klamkin, Tanneau, Van Hentenryck}
\begin{document}
\maketitle

\begin{abstract}
\small

In many practical applications of constrained optimization, scale and
solving time limits make traditional optimization solvers
prohibitively slow.  Thus, the research question of how to design
optimization proxies -- machine learning models that produce
high-quality solutions -- has recently received significant attention.
Orthogonal to this research thread which focuses on learning primal
solutions, this paper studies how to learn dual feasible solutions
that complement primal approaches and provide quality guarantees.  The
paper makes two distinct contributions. First, to train dual linear
optimization proxies, the paper proposes a smoothed self-supervised
loss function that augments the objective function with a dual penalty
term. Second, the paper proposes a novel dual completion strategy that
guarantees dual feasibility by solving a convex optimization
problem. Moreover, the paper derives closed-form solutions to this
completion optimization for several classes of dual penalties,
eliminating the need for computationally-heavy implicit layers.
Numerical results are presented on large linear optimization problems
and demonstrate the effectiveness of the proposed approach.
The proposed dual completion outperforms
methods for learning optimization proxies which do not
exploit the structure of the dual problem.  Compared to
commercial optimization solvers, the learned dual proxies 
achieve optimality gaps below $1\%$
and several orders of magnitude speedups.

\end{abstract}

\section*{Keywords}
Machine learning, linear optimization, duality, multi-parametric optimization, interior point method

\section{Introduction}
\label{sec:intro}

Optimization technology is routinely used to plan and operate large-scale systems in virtually all sectors of the economy, 
from bulk power systems operations to production planning, logistics and supply chains management \cite{Lodi2010}.
Many of these problems are solved repeatedly, e.g., on a daily or hourly basis, with only small changes in problem structure or input data.
This has spurred significant interest in data-driven methods for real-time decision making, 
which alleviate the computational bottlenecks of traditional optimization tools 
by leveraging abundant historical data and advanced Machine Learning (ML) architectures.

Substantial progress has been achieved in developing so-called constrained optimization ``proxy'' models
that learn the mapping from instance data to an optimal solution \cite{Bengio2021_ML4CO,kotary2021end}.
Once trained, optimization proxies can produce high-quality solutions in milliseconds.
Most prior work in constrained optimization learning focuses on learning primal solutions,
with considerable effort dedicated towards guaranteeing the feasibility of predicted solutions, see,
e.g., the large body of work on learning proxies for Optimal Power Flow (OPF) problems \cite{khaloie2024review}.
Another important line of work is the design of self-supervised learning algorithms,
which improve training efficiency by eliminating the need for labeled data \cite{donti2021dc3,Park2023_PDL}, and have been shown to achieve better performance 
than supervised models \cite{e2efeasible}.

A fundamental limitation of ML-based optimization proxies
is their lack of formal performance guarantees.
Recent work addresses this limitation by learning \emph{dual} proxies \cite{qiu2023dual,Demelas2024_PredictingLagrangianMIP,tanneau2024dual}
that predict dual-feasible solutions, thus providing valid dual bounds.
Building on this approach, this paper improves the training of dual optimization proxies 
by augmenting the training loss function with a barrier-based regularizer.
While barrier-based smoothing has been used in a decision-focused learning setting \cite{mandi2020interior},
this is the first paper to propose it for learning optimization proxies.
In addition to this new mathematical framework, the paper presents closed-form, analytical formulae
for efficient forward and backward passes, thus improving training and inference times by several
orders of magnitude compared to implicit differentiation methods \cite{agrawal2019differentiating}.

\subsection{Contributions and Outline}

The paper makes the following core contributions.
First, the paper proposes a smoothed self-supervised learning (S3L) loss function for training
dual optimization proxies, using a regularization mechanism inspired from interior-point algorithms.
Second, it presents closed-form, analytical formulae for efficient forward and backward passes,
thus improving training and inference times by several
orders of magnitude compared to implicit differentiation methods.
Third, it evaluates the practical performance of the
proposed methods and compares them to state-of-the-art
techniques on large-scale parametric optimal power flow problems.

The rest of the paper is organized as follows.
Section \ref{sec:intro:notations} introduces relevant notations.
Section \ref{sec:S3L} presents the paper's proposed S3L methodology.
Section \ref{sec:experiments} reports numerical results, 
and Section \ref{sec:conclusion} concludes the paper.

\subsection{Notations}
\label{sec:intro:notations}

Bold lowercase letters refer to vectors and bold uppercase letters
refer to matrices. Scalars are denoted by non-bold letters.  
The $\odot$ symbol represents element-wise multiplication.
A vector in the denominator of a fraction represents element-wise division.
Similarly, exponents on vectors represent element-wise exponentiation.

The positive and negative parts of $x \, {\in} \, \mathbb{R}$ are denoted by $|x|^{+} \, {=} \, \max(0, x)$ and $|x|^{-} \, {=} \, \max(0, -x)$.
Note that $|x|^{+}, |x|^{-} \, {\geq} \, 0$ and that $x = |x|^{+} \, {-} \, |x|^{-}$.
For $\bx \, {\in} \, \mathbb{R}^{n}$, $|\bx|^{+}, |\bx|^{-}$ denote the element-wise positive and negative parts of $\bx$, respectively.

The paper considers multi-parametric linear programming (MPLP) problems, stated in primal-dual form\\
\begin{minipage}{0.4\textwidth}
\begin{subequations}
    \label{model:primal}
    \begin{align}
        (\textsc{P}_{\beta}) \quad 
        \min_\bx \quad 
            & \bc_\beta^\top \bx  \\
        \text{s.t.} \quad 
            & \bA_\beta\bx = \bb_\beta \\
            & \bl_\beta \leq \bx \leq \bu_\beta \label{model:primal:bound}
    \end{align}
\end{subequations}
\end{minipage}
\hfill
\begin{minipage}{0.59\textwidth}
\begin{subequations}
    \label{model:dual}
    \begin{align}
        (\textsc{D}_{\beta}) \quad
        \max_{\by, \bzl, \bzu} \quad
            & \label{model:dual:obj}
            \bb_\beta^{\top}\by + \bl_\beta^{\top}\bzl - \bu_\beta^{\top}\bzu\\
        \text{s.t.} \quad 
            & \bA_\beta^{\top}\by + \bzl - \bzu = \bc_\beta\label{model:dual:eqconstr}\\
            & \bzl, \,\bzu \geq 0\label{model:dual:bound}
    \end{align}
\end{subequations}
\end{minipage}\\
where $\beta \in \mathbb{R}^{p}$ denotes the instance parameters, $\bc_\beta,\,\bl_\beta,\,\bu_\beta,\,\bx,\,\bzl,\,\bzu \in\mathbb{R}^n$,
$\bb_\beta,\,\by \in\mathbb{R}^m$, and $\bA_\beta\in\mathbb{R}^{m \times n}$.
The paper assumes that $(\textsc{P}_{\beta})$ and $(\textsc{D}_{\beta})$ are always feasible,
hence, both are solvable and strong duality always holds.
In addition, the variable bounds \eqref{model:primal:bound} are non-trivial and consistent, i.e.,
$\bl_{\beta} < \bu_{\beta}$.
The feasible set of $(\textsc{D}_{\beta})$ is denoted by $\Lambda(\beta)$, i.e., 
$\Lambda(\beta) = \{ (\by, \bzl, \bzu) | \eqref{model:dual:eqconstr}, \eqref{model:dual:bound} \}$.
Finally, for brevity, the subscript $\beta$ is dropped in the rest of the paper.
\section{Smoothed Self-Supervised Learning for Dual Optimization Proxies}
\label{sec:S3L}

\begin{figure}[t!]
    \color{black}
    \centering
    \includegraphics[width=0.9\textwidth]{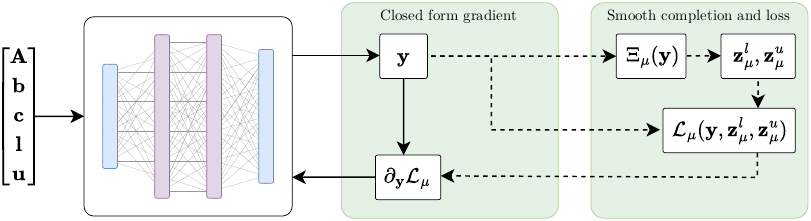}
    \vspace{-0.5em}
    \caption{Illustration of the proposed S3L method.
        From left to right: given input data $(\bA, \bb, \bc, \bl, \bu$), a neural network first predicts $\by$.
        A generalized dual completion layer then recovers $(\bzl_{\mu}, \bzu_{\mu})$ in closed form (Section \ref{sec:forward}), and the loss $\mathcal{L}_\mu(\by,\bz^l_\mu,\bz^u_\mu)$ is evaluated.
        Gradient information $\partial_{\by} \mathcal{L}_\mu$ can be computed either by automatic differentiation (following operations marked with dashed arrows) or in closed-form from $\by$ directly (Section \ref{sec:backward}).
    }
    \label{fig:completiontikz}
    \vspace{-1em}
\end{figure}

This section presents the proposed S3L methodology, which is illustrated in Figure \ref{fig:completiontikz}.
The method's core methodological contributions comprise 
1) a smoothed loss for self-supervised training of
dual linear optimization proxies, and 
2) closed-form analytical formulae
for efficient forward and backward passes.

Following \cite{tanneau2024dual}, the self-supervised training of a dual optimization proxy is stated as the optimization problem
    \begin{equation}
    \label{eq:dual_proxy_training}
    \begin{aligned}
        \max_{\theta} \quad 
        &
        \mathbb{E}_{\beta \sim \mathcal{B}} 
        \left[
            \bb^{\top} \by + \bl^{\top} \bzl - \bu^{\top} \bzu
        \right]\\
        \text{s.t.} \quad
        &
        (\by, \bzl, \bzu) = \mathcal{F}_{\theta}(\beta)\\
        &
        (\by, \bzl, \bzu) \in \Lambda(\beta), \quad \forall \beta \sim \mathcal{B}.
    \end{aligned}
    \end{equation}
    which seeks to maximize the expected (dual) objective value of the predicted dual solution $(\by, \bzl, \bzu) = \mathcal{F}(\beta)$, subject to dual feasibility constraints.
    Therein, $\mathcal{F}_{\theta}$ denotes an artificial neural network (NN) with weights $\theta$.
    When solving \eqref{eq:dual_proxy_training} with a stochastic gradient-based algorithm, each step mimics that of a dual (sub)gradient ascent algorithm \cite{boyd2004convex}, which is known to exhibit slow convergence.
    To alleviate this issue, the paper proposes to \emph{smooth} the training loss, by augmenting it with a regularizing term.
    Namely, the proposed smoothed self-supervised learning (S3L) loss is
    \begin{align}
        \label{eq:S3L}
        \mathcal{L}_{\mu}(\by, \bzl, \bzu) = \bb^{\top}\by + \bl^{\top} \bzl - \bu^{\top}\bzu + \mu \cdot \Phi(\by, \bzl, \bzu),
    \end{align}
    where $\Phi$ is concave and $\mu {\geq} 0$ controls the weight of the regularization.
    When considering a single instance, the training problem is then akin to solving the smoothed dual problem
    \begin{align}
        \label{eq:smooth_dual}
        \max_{\by, \bzl, \bzu} \quad 
        \left\{ 
            \bb^{\top}\by + \bl^{\top} \bzl - \bu^{\top}\bzu + \mu \cdot \Phi(\by, \bzl, \bzu)
            \ \middle| \ 
            \eqref{model:dual:eqconstr}, \eqref{model:dual:bound} 
        \right\}.
    \end{align}
    The use of a regularization term to smooth the dual problem is well-known in the optimization literature, see, e.g., \cite{Nesterov2005_SmoothMinimization} for a general treatment of how to optimize non-smooth functions.
    To the authors' knowledge, this paper is the first to propose dual smoothing for learning optimization proxies.

    The proposed S3L yields a generalized dual completion strategy where,
    given $\by \in \mathbb{R}^{m}$, variables $(\bzl_{\mu}, \bzu_{\mu})$ are obtained as the solution of
    \begin{subequations}
        \label{eq:smooth_inner_problem}
        \begin{align}
            \Xi_{\mu}(\by) = \max_{\bzl, \bzu\geq0} \quad
                & \bl^{\top} \bzl - \bu^{\top} \bzu + \mu \cdot \Phi(\by, \bzl, \bzu)\\
            \text{s.t.} \quad
                & \bzl - \bzu = \bc - \bA^{\top} \by.  \label{eq:smooth_inner_problem:con:eq}
        \end{align}
    \end{subequations}
    In general, embedding \eqref{eq:smooth_inner_problem} in a neural network architecture requires the use
    of differentiable optimization layers \cite{agrawal2019differentiable}, which is computationally costly and numerically unstable.
    To alleviate this issue, the paper presents closed-form analytical formulae for the case where
    $\Phi$ is a logarithmic barrier function.
    In all that follows, assume that $\Phi$ is of the form
    $\Phi(\by, \bzl, \bzu) = \sum_{j=1}^{n} \varphi(\bzl_{j}) + \varphi(\bzu_{j})$
    where $\varphi$ is a strictly concave scalar function. 
    Specific results are presented next for the logarithmic barrier $\varphi(t){=}\ln(t)$, which is at the core of interior-point optimization algorithms \cite{boyd2004convex}.

\subsection{S3L: the Forward Pass}
\label{sec:forward}

    A key result of the paper is that the inner problem \eqref{eq:smooth_inner_problem} admits closed-form solutions for both the unregularized case and the logarithmic barrier, which are presented in Theorems \ref{thm:completion:noreg} and \ref{thm:completion:log} respectively.
    Closed-form solutions eliminate the need for implicit optimization layers during training and inference,
    which offers substantial computational benefits.

        \begin{theoremrep}
            \label{thm:completion:noreg}
            Assume $\mu \, {=} \, 0$, let $\by \, {\in} \, \mathbb{R}^{m}$, and define $\bz \, = \, \bc {-} \bA^{\top} \by$.
            Then,
                   $\bzl_{\mu}(\by) = |\bz|^{+}$,
                   and
                   $\text{ }\bzu_{\mu}(\by) = |\bz|^{-}.$
        \end{theoremrep}
        \begin{appendixproof}
            Direct application of \cite[Example 1]{tanneau2024dual}.
        \end{appendixproof}

        \begin{theoremrep}
            \label{thm:completion:log}
            Assume $\mu \, {>} \, 0$ and $\varphi(t) \, {=} \, \ln(t)$.
            Let $\by \, {\in} \, \mathbb{R}^{m}$, and define $\bz \, {=} \, \bc {-} \bA^{\top} \by$.
            Then,
                \begin{align}
                \label{eq:completion:log}
                    \bzl_{\mu}(\by) = \mathbf{v} + \mathbf{w} + \sqrt{\mathbf{v}^{2} + \mathbf{w}^{2}}
                    \quad \text{ and } \quad
                    \bzl_{\mu}(\by) = \mathbf{v} - \mathbf{w} + \sqrt{\mathbf{v}^{2} + \mathbf{w}^{2}},
                \end{align}
                where $\mathbf{v} = \mu / (\bu - \bl)$, $\mathbf{w} = \bz / 2$, and all operations are element-wise.
        \end{theoremrep}
        \begin{appendixproof}
            The KKT conditions of the regularized inner dual problem are
                    $$\quad\bzl - \bzu = \bz, \quad\quad
                    \bl \leq \bx \leq \bu, \quad\quad
                    ((\bx - \bl)\odot \bzl)_i = \mu, \quad\quad
                    ((\bu - \bx)\odot \bzu)_i = \mu.$$
            It is straightforward to verify that \eqref{eq:completion:log} with $\bx_{\mu}(\by)$ from Theorem \ref{thm:backward:log} satisfy the KKT equations above.
            Therefore, \eqref{eq:completion:log} is optimal for \eqref{eq:smooth_inner_problem}, and $\bx_{\mu}(\by)$ is an optimal dual solution associated with constraints \eqref{eq:smooth_inner_problem:con:eq}.
        \end{appendixproof}
        
\subsection{S3L: the Backward Pass}
\label{sec:backward}

During training, one must compute gradients of the loss function with respect the NN weights $\theta$.
This is typically done by applying the chain rule layer by layer, also known as back-propagation.
This section presents closed-form gradients for the S3L loss and dual completion layer, providing an alternative to automatic differentiation with better numerical stability.
A first general result on the form the gradients of $\mathcal{L}_{\mu}$ is given by Theorem \ref{thm:backward:theory}.

    \begin{theoremrep}
        \label{thm:backward:theory}
        Assume $\mu \, {\geq} \, 0$.
        Then, $\Xi_{\mu}$ is concave and  admits supergradients of the form $\partial_{\by} \Xi_{\mu}(\by) \, {=} \, {-} \bA \bx_{\mu}(\by)$, where $\bx_{\mu}(\by)$ is a dual-optimal solution associated with constraint \eqref{eq:smooth_inner_problem:con:eq}.
        If, in addition, $\mu > 0$ and $\varphi$ is strictly concave, then $\Xi_{\mu}$ is strictly concave and differentiable.
    \end{theoremrep}
    \begin{appendixproof}
        The concavity of $\Xi_{\mu}$ stems from the fact that $\Xi_{\mu}(\by)$ is the value function of \eqref{eq:smooth_inner_problem}, which is a concave maximization problem whose right-hand side is an affine function of $\by$ \cite{griva2008linear}.
        The relation $\nabla_{\by}\Xi_{\mu}(\by) = -\bA \bx_{\mu}(\by)$ holds by definition of dual variables \cite[Section 14]{griva2008linear}.
    \end{appendixproof}
    
    Noting that $\mathcal{L}_{\mu}(\by, \bzl_{\mu}, \bzu_{\mu}) = \bb^{\top} \by + \Xi_{\mu}(\by)$,
    it then follows that (sub)gradients $\partial_{\by} \mathcal{L}_{\mu}(\by)$ are of the form $\bb \, {-} \, \bA \bx_{\mu}(\by)$.
    This relation allows to derive efficient formulae to compute $\partial_{\by} \mathcal{L}_{\mu}(\by)$ without requiring implicit layers nor auto-differentiation tools, which can improve performance and numerical stability.
    Theorem \ref{thm:backward:theory} further shows that the smoothed loss is differentiable, in contrast to the unregularized case, which only admits sub-gradients.

    Next, Theorem \ref{thm:backward:non} and Theorem \ref{thm:backward:log} present closed-form analytical formulae
    for computing $\bx_{\mu}$ in the unregularized and logarithmic barrier cases, respectively.
    Such closed-form expressions allow for more efficient gradient computations compared to automatic differentiation,
    and have been found to be more numerically stable.
    
        \begin{theoremrep}
        \label{thm:backward:non}
            Assume $\mu \, {=} \, 0$, and let $\by \, {\in} \, \mathbb{R}^{m}$.
            A dual-optimal solution to \eqref{eq:smooth_inner_problem} is given by
        \begin{align}
            \label{eq:ss-yopt}
            (\bx_{\mu}(\by))_{j} = \bl_{j} \enspace\text{if}\enspace  \bz_j > 0,\quad (\bx_{\mu}(\by))_{j} = \bu_{j} \enspace\text{if}\enspace  \bz_j < 0, \quad\text{and}\enspace \hat{\bx}_{j} \in [\bl_{j}, \bu_{j}] \enspace\text{if}\enspace \bz_j=0.
        \end{align}
        \end{theoremrep}
        \begin{appendixproof}
            See proof of Theorem \ref{thm:completion:noreg}; the optimality of $\bx_{0}(\by)$ is obtained from KKT conditions.
        \end{appendixproof}

        \begin{theoremrep}
            \label{thm:backward:log}
            Assume $\mu > 0$ and $\varphi(t) = \ln(t)$. Let $\by \in \mathbb{R}^{m}$, and define $\bz = \bc - \bA^{\top} \by$.
            A dual-optimal solution to \eqref{eq:smooth_inner_problem} is given by
            $
                \left(\bx_{\mu}(\by) \right)_j 
                = 
                \frac{\bu_j+\bl_j}{2} +
                     \frac{\mu}{\bz_j} -\operatorname{sign}\big(\bz_j\big)\sqrt{(\frac{\mu}{\bz_j})^2+(\frac{\bu_j-\bl_j}{2})^2}
            $ if $\bz_{j} \neq 0$, 
            and $\left(\bx_{\mu}(\by) \right)_j = (\bu_j+\bl_j)/{2}$ if $\bz_{j} = 0$.
            
        \end{theoremrep}
        \begin{appendixproof}
            See proof of Theorem \ref{thm:completion:log}; the optimality of $\bx_{\mu}(\by)$ is obtained from KKT conditions.
        \end{appendixproof}

\newcommand{\penloss}{\textsc{Penalty}}

\section{Numerical Experiments}
\label{sec:experiments}

\subsection{Optimal Power Flow Formulation}
\label{sec:experiments:OPF}

Consider parametric DC optimal power flow (DCOPF) problems of the form
\begin{subequations}
    \label{eq:DCOPF}
    \begin{align}
        \text{DCOPF}(\pd) \quad \quad 
        \min_{\pg,\, \pf} \quad & \mathbf{c}^{\top} \pg \\
        \text{s.t.} \quad
            & \mathbf{e}^{\top} \pg = \mathbf{e}^{\top} \pd_\beta,\\
            & \pf = \PTDF (\pg - \pd_\beta),\\
            & \pgmin \leq \pg \leq \pgmax,\\
            & \pfmin \leq \pf \leq \pfmax,
    \end{align}
\end{subequations}
where variables $\pg$ and $\pf$ denote generation dispatch and power flows, respectively, 
and $\pd_{\beta}$ denotes the parametric energy demand, which varies across instances.
The formulation uses the industry-standard Power Transfer Distribution Factor (PTDF)
formulation to express power flows ($\pf$) as a function of generation ($\pg$) and demand ($\pd_{\beta}$).

The numerical experiments consider large-scale DCOPF instances from PGLib \cite{pglib},
which originate from European high-voltage transmission grids:
\texttt{1354\_pegase}, \texttt{2869\_pegase},
\texttt{6470\_rte}, and \texttt{9241\_pegase}.
For the largest system (\texttt{9241\_pegase}), Problem \eqref{eq:DCOPF} comprises over 10,000 variables and constraints,
about two orders of magnitude larger than the instances considered in \cite{tanneau2024dual}.
DCOPF instances are made parametric by varying the demand ($\pd$),
using the same data generation methodology as \cite[Section 3]{pglearn}.

\subsection{Experiment details}

The proposed S3L is evaluated against the following approaches:
Dual Lagrangian Learning (DLL) \cite{tanneau2024dual}, which corresponds to setting $\mu = 0$,
Deep Constraint Completion and Correction (DC3) \cite{donti2021dc3},
and a vanilla neural network wherein constraint violations are penalized during training (\penloss).
All methods use a similar underlying neural network architecture, and are trained in a self-supervised fashion, i.e., they do not require access to optimal solutions.
Hence, the methods only differ in their completion mechanism and loss function using during training.
In S3L and DLL, the NN predicts $\by$, and $(\bzl, \bzu)$ are obtained by dual completion (see Section \ref{sec:forward}).
Both methods use the loss function $\mathcal{L}_{\mu}$ as in \eqref{eq:S3L}; recall that DLL corresponds to $\mu=0$.
In DC3, the NN predicts $(\by, \bzl)$, and $\bzu$ is recovered using \eqref{model:dual:eqconstr}.
This completion step is followed by an iterative correction mechanism \cite{donti2021dc3}.
In \penloss, the NN predicts $(\by, \bzl, \bzu)$, and no completion nor correction is performed.
The NN architectures in DC3 and {\penloss} use softplus activations to ensure $\bzl \geq 0$ and $\bzl, \bzu \geq 0$, respectively.
The training loss for DC3 and {\penloss} maximizes the dual objective value, and penalizes violations of \eqref{model:dual:eqconstr}-\eqref{model:dual:bound};
note that, unlike the completion mechanism used in S3L and DLL, the solutions produced by DC3 and {\penloss} may be infeasible.

In all experiments, the NN architecture consists of a 4-layer fully-connected layers with a hidden layer size of 128 and softplus activations.
All ML models are implemented with PyTorch 2.5.1 \cite{pytorch} and run on a single NVIDIA L40S GPU.
LP instances are solved using HiGHS \cite{highs}.
Each dataset comprises 40,000 samples; $80\%$ is used for training and the remaining $20\%$ for testing.
Recall that optimal solutions are not seen during training, and are only used to evaluate each method's performance on the test set.
Finally, each experiment is repeated 10 times with different seeds, to study how robust the methods are to different training/testing splits and neural network weight initialization.

For {\textsc{DC3}}, following \citet{donti2021dc3}, 
the momentum is set to 0.5, the number of correction steps to 10, and the step size to $10^{-2}$.
For both {\textsc{DC3}} and {\penloss}, the violation weight is set to $10^8$, about $100$ times the mean dual objective value.
Finally, in {\textsc{S3L}}, the barrier penalty parameter $\mu$ is initialized to $\mu^{(0)} = 1$,
and reduced at the end of each epoch $k$ following an exponential decay scheme, namely, $\mu^{(k+1)} \leftarrow 0.99 \mu^{(k)}$.

\subsection{Results}
\label{sec:results}

Each method's performance is evaluated using the following metrics, where $(\by, \bzl, \bzu)$ denote the predicted dual solution.
Dual constraint violations are reported as $\mathcal{V} = |\bzl|^{-} + |\bzu|^{-} + |\bA^\top\by + \bzl -\bzu -\bc|$.
The objective difference $\Delta \mathcal{G} = {|\mathcal{L}_\star - (\bb^{\top} \by + \bl^{\top} \bzu - \bu^{\top}\bzu)|}\;/\,{\mathcal{L}_\star}$ measures the relative difference in objective value between the predicted solution, and the optimal value $\mathcal{L}_{\star}$.
Note that, if $(\by, \bzl, \bzu)$ is infeasible, then $(\bb^{\top} \by + \bl^{\top} \bzu - \bu^{\top}\bzu)$ is not a valid dual bound.
Therefore, to ensure a fair comparison across all methods, the paper also reports the dual optimality gap $\mathcal{G}_{\star} = {|\mathcal{L}_\star - (\bb^{\top} \by + \Xi_{0}(\by))|}\;/\,{\mathcal{L}_\star}$, which uses the dual completion of Theorem \ref{thm:completion:noreg} to ensure dual feasibility and, hence, valid dual bounds.
Naturally, if $(\by, \bzl, \bzu)$ is feasible, then $\Delta \mathcal{G} = \mathcal{G}_{\star}$.
Finally, to measure each methods' robustness, the paper reports the worst dual gap, denoted by $\mathcal{G}^{\max}_{\star}$ and defined as the maximum value taken by $\mathcal{G}_{\star}$ on the test set.
All three metrics $\Delta \mathcal{G}$, $\mathcal{G}_{\star}$ and $\mathcal{G}^{\max}_{\star}$ are expressed as a percentage.

\begin{figure}
    \centering
    \includegraphics[width=\textwidth]{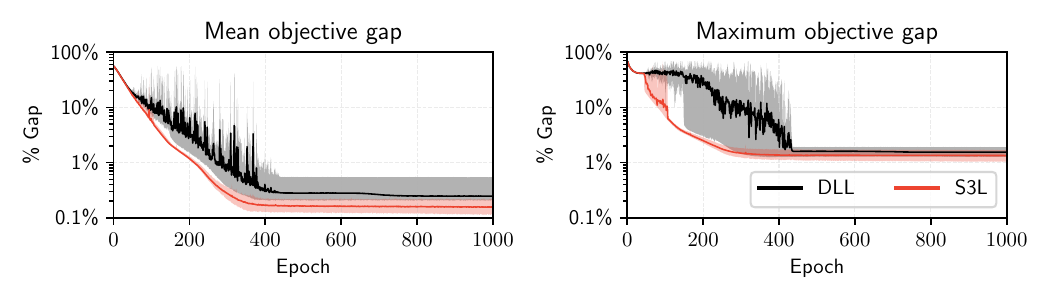}
    \vspace{-2em}
    \caption{Training curves for DLL and S3L on the \texttt{2869\_pegase} case.
        Left: evolution of mean objective gap.
        Right: evolution of maximum objective gap.
        The shaded area indicates the range between best and worst seed.}
    \label{fig:2869_pegase}
\end{figure}

Table \ref{tab:bperformance} compares the performance of each method on the four datasets.
Overall, {\penloss} displays the largest violations, which is expected given its absence of completion step.
Next, although DC3's correction step achieves zero violations on all but the largest dataset (\texttt{9241\_pegase}), the method fails to produce meaningful dual bounds across all datasets,
with average gaps over 1000\% and worst gaps above 2000\%.
The poor performance of DC3 is attributed to its numerical difficulties, which have been observed in other dual learning settings \cite{qiu2023dual,tanneau2024dual}.
In contrast, DLL and S3L both achieve dual gaps below 0.5\% on average, and never higher than about 1.5\%; this performance can be attributed to the use of a dual completion layer.
Furthermore, S3L outperforms DLL on all datasets, both on average and in the worst case, with a reduction in average gap of 10\% to 25\%.
Learning curves displayed in Figure \ref{fig:2869_pegase} confirm the improved performance of S3L,
which displays a steady progress both on average and in the worst case.

Finally, a comparison of computing times confirms that ML-based methods outperform traditional optimization solvers by about two orders of magnitude.
On the largest system, {\penloss}, DC3, DLL and S3L can process 1,000 instances in about 25ms, 80ms, 22ms and 23ms on a single GPU, respectively, whereas HiGHS requires 1200ms on 32 CPU cores.
The slower performance of DC3 is caused by its iterative correction step.
In contrast, DLL and S3L use a closed-form dual completion step, with the added benefit of guaranteed dual feasibility.

\begin{table}[!t]
\renewcommand{\arraystretch}{1}%
\centering
\tabcolsep = 0.45\tabcolsep
    \centering
        \caption{Performance comparison on parametric DCOPF cases.}
        \begin{tabular}{l@{\hskip1.25\tabcolsep}rrrr@{\hskip1.25\tabcolsep}rrrr@{\hskip1.25\tabcolsep}rrrr@{\hskip2\tabcolsep}rrrr}
        \toprule
                              & \multicolumn{4}{c}{\texttt{1354\_pegase}}  & \multicolumn{4}{c}{\texttt{2869\_pegase}} & \multicolumn{4}{c}{\texttt{6470\_rte}} & \multicolumn{4}{c}{\texttt{9241\_pegase}}                \\[-0.2\normalbaselineskip]
         \cmidrule(lr){2-5} \cmidrule(lr){6-9} \cmidrule(lr){10-13} \cmidrule(lr){14-17} \\[-1.125\normalbaselineskip]
         {}                     & \multicolumn{1}{c}{$\mathcal{G}_\star$} & \multicolumn{1}{c}{$\mathcal{G}_\star^{\max}$} & \multicolumn{1}{c}{$\mathcal{V}$}  & \multicolumn{1}{c}{$\Delta\mathcal{G}$}
                                & \multicolumn{1}{c}{$\mathcal{G}_\star$} & \multicolumn{1}{c}{$\mathcal{G}_\star^{\max}$} & \multicolumn{1}{c}{$\mathcal{V}$}  & \multicolumn{1}{c}{$\Delta\mathcal{G}$}
                                & \multicolumn{1}{c}{$\mathcal{G}_\star$} & \multicolumn{1}{c}{$\mathcal{G}_\star^{\max}$} & \multicolumn{1}{c}{$\mathcal{V}$}  & \multicolumn{1}{c}{$\Delta\mathcal{G}$}
                                & \multicolumn{1}{c}{$\mathcal{G}_\star$} & \multicolumn{1}{c}{$\mathcal{G}_\star^{\max}$} & \multicolumn{1}{c}{$\mathcal{V}$}  & \multicolumn{1}{c}{$\Delta\mathcal{G}$}\\
        \midrule
        \textsc{Penalty}
                         & \multicolumn{1}{|r}{1119}  & 1522 & 1299  & 71.1 
                         & \multicolumn{1}{|r}{1134}  & 1595 & 2920  & 61.1 
                         & \multicolumn{1}{|r}{3487}  & 4905 & 6897  & 82.8 
                         & \multicolumn{1}{|r}{1394}  & 1952 & 7613  & {52.9}     \\
        \textsc{DC3}
                         & \multicolumn{1}{|r}{1656}  & 2525 & 0.00  & 1697 
                         & \multicolumn{1}{|r}{1285}  & 2052 & 0.00  & 1331 
                         & \multicolumn{1}{|r}{1173} & 2012 & 0.00  & 1263 
                         & \multicolumn{1}{|r}{1384}  & 2121 & 277  & 1461     \\
        \midrule
        \textsc{DLL}
                         & \multicolumn{1}{|r}{0.25}  & 0.57 & 0  & 0.25 
                         & \multicolumn{1}{|r}{0.26}  & 1.55 & 0  & 0.26 
                         & \multicolumn{1}{|r}{0.31}  & 0.84 & 0  & 0.31 
                         & \multicolumn{1}{|r}{0.20}  & 0.91 & 0  & {0.20}     \\
        $\textsc{S3L}$
                         & \multicolumn{1}{|r}{\textbf{0.17} } & \textbf{0.38} & 0  & \textbf{0.17} 
                         & \multicolumn{1}{|r}{\textbf{0.16}}  & \textbf{1.34} & 0  & \textbf{0.16} 
                         & \multicolumn{1}{|r}{\textbf{0.27}}  & \textbf{0.76} & 0  & \textbf{0.27} 
                         & \multicolumn{1}{|r}{\textbf{0.16}}  & \textbf{0.84} & 0  & {\textbf{0.16}}   \\
        \bottomrule
    \end{tabular}
    \vspace{1em}
    \label{tab:bperformance}
\end{table}

\section{Conclusion}
\label{sec:conclusion}

This paper presented S3L, a new methodology to design dual proxies for multi-parametric  bounded linear programs. 
Its main innovations are (1) a smooth self-supervised learning (S3L) loss function for training dual optimization proxies; and (2) closed-form dual completion layers and gradients for speeding up the forward and backward passes, thus enabling efficient training.
The proposed generalized dual completion layers extend previous work \cite{tanneau2024dual} by explicitly taking into account a regularizer term in the loss function.
The resulting S3L architecture is computationally efficient and exhibits faster and more robust progress during training.
The proposed method was evaluated on industry-size optimal power flow problems to demonstrate its performance.
Numerical results show that S3L achieves state-of-the-art performance in both accuracy and speed, improving dual gaps by 10--25\% compared to previous work, while maintaining a 500x speedup over a leading simplex-based LP solver.

\section*{Acknowledgements}
This research was partly funded by NSF award 2112533, ARPA-E PERFORM award AR0001136, and NSF award DGE-2039655. Any opinions, findings, and conclusions or recommendations expressed in this material are those of the author(s) and do not necessarily reflect the views of NSF or ARPA-E.

\bibliography{bibliography}

\end{document}